\documentclass[conference]{IEEEtran}
\usepackage{cite}
\usepackage{amsmath,amssymb,amsfonts}
\usepackage{graphicx}
\usepackage{textcomp}
\usepackage{xcolor}
\usepackage{fancyhdr}
\usepackage{txfonts}
\usepackage{algorithm}
\usepackage{algpseudocode}
\floatname{algorithm}{Procedure}

\def\BibTeX{{\rm B\kern-.05em{\sc i\kern-.025em b}\kern-.08em
    T\kern-.1667em\lower.7ex\hbox{E}\kern-.125emX}}

\fancypagestyle{FirstPage}{

\cfoot{\fontsize{9pt}{11pt}\selectfont 
    \textit{
        \textcopyright 2022 IEEE. Personal use of this material is permitted. Permission
        from IEEE must be obtained for all other uses, in any current or future
        media, including reprinting/republishing this material for advertising or
        promotional purposes, creating new collective works, for resale or
        redistribution to servers or lists, or reuse of any copyrighted
        component of this work in other works.
        }
    }
}
    
\begin{document}

\title{Co-evolving morphology and control of soft robots using a single genome}

\author{
\IEEEauthorblockN{Fabio Tanaka}
\IEEEauthorblockA{\textit{Graduate School of Science and Technology} \\
\textit{University of Tsukuba}\\
Tsukuba, Japan \\
tanaka.fabio.xa@alumni.tsukuba.ac.jp}
\and
\IEEEauthorblockN{Claus Aranha}
\IEEEauthorblockA{\textit{Department of Computer Sciences} \\
\textit{University of Tsukuba}\\
Tsukuba, Japan \\
caranha@cs.tsukuba.ac.jp}}

\maketitle

\begin{abstract}
When simulating soft robots, both their morphology and their controllers play important roles in task performance. This paper introduces a new method to co-evolve these two components in the same process. We do that by using the hyperNEAT algorithm to generate two separate neural networks in one pass, one responsible for the design of the robot body structure and the other for the control of the robot.

The key difference between our method and most existing approaches is that it does not treat the development of the morphology and the controller as separate processes. Similar to nature, our method derives both the ``brain" and the ``body" of an agent from a single genome and develops them together. While our approach is more realistic and doesn't require an arbitrary separation of processes during evolution, it also makes the problem more complex because the search space for this single genome becomes larger and any mutation to the genome affects ``brain" and the ``body" at the same time.

Additionally, we present a new speciation function that takes into consideration both the genotypic distance, as is the standard for NEAT, and the similarity between robot bodies. By using this function, agents with very different bodies are more likely to be in different species, this allows robots with different morphologies to have more specialized controllers since they won't crossover with other robots that are too different from them.

We evaluate the presented methods on four tasks and observe that even if the search space was larger, having a single genome makes the evolution process converge faster when compared to having separated genomes for body and control. The agents in our population also show morphologies with a high degree of regularity and controllers capable of coordinating the voxels to produce the necessary movements.

\end{abstract}


\section{Introduction}
\thispagestyle{FirstPage}
The evolution of body structures of simulated agents, as well as the means to control that body to achieve a specific goal, is a topic of interest in Artificial Life. In nature, creatures occupying different niches evolved different body types, and are able to freely control them to guarantee their survival.

However, simulating the evolution of the body and control simultaneously is a complex problem since distinct bodies may need different control algorithms to function. For example, a creature without legs that crawls on the ground moves differently from a quadruped creature that moves by stepping on its legs.

One common approach for this challenge is to consider the evolution of the morphology and control as separate processes. Some previous works focused on evolving the controller for a fixed body \cite{self-attention}. Others alternated between the optimization of robot morphology and robot control. \cite{evogym}.

In this study, we co-evolve morphology and controller in the same process. While this approach presents some challenges, namely the increased chance of destructive mutations (because changes on the body may affect the controller and vice-versa) and the larger search space (because now we have to derive two outcomes from a single genome); we believe this leads to a more realistic simulation because, in nature, the development of body and brain occurs as a single process and is derived from the same genome.  

We represent the controller and morphology using an Artificial Neural Networks (ANNs) . To express both ANNs by a single genome, we utilize the idea behind the hyperNEAT algorithm, where a Compositional Pattern-Producing Network (CPPN) generates both ANNs. 

To simulate the creatures we use 2D Voxel-based Soft Robots (VSRs). This allows the agent to build its own body by combining different types of blocks, also called voxels, and control them by varying their volume in response to a control signal.

\section{Related Work}
In this research, we focus on the problem of co-evolving the body (i.e. structure or morphology) and control (i.e. brain) of soft robots. Most works in the literature approach this problem by considering the evolution of the body and the control as separate processes. 

An example of a study that focuses on developing the brain while having the morphology fixed is the research by Pigozzi, Tang, Medvet, and Ha~\cite{self-attention}, where they evolved distributed controllers for VSRs. Their robots did not have a central control system but had, in each voxel, an ANN that output signals to adjacent blocks. The ANN also controls the volume of its voxel based on these signals and local sensors. They tested different body structures but fixed them while evolving the robots.

On the other side of the spectrum, Cheney, MacCurdy, Clune, and Lipson \cite{unshackling} chose to focus on the development of the VSR morphology. They used a Compositional Pattern-Producing Network (CPPN) to decide the presence and type of voxels in a 3D VSR environment. There was no central control system in their study, their voxels would expand or contract at fixed frequencies that were set at the beginning of the experiments.

Finally, studies aiming to develop both body and control are usually formulated as a two-level optimization problem. This approach involves a design optimization method that evolves the physical configurations of the robots in the outer loop and a control optimization algorithm that computes an optimized controller for a given robot body in the inner loop. Reference \cite{evogym} is an example of this type of work; in it, the body structure is evolved by using the NEAT algorithm, and the controller is developed by using PPO optimization.

Although these approaches produce agents that can efficiently solve tasks, they consider the body and the controller of the robots as separate entities that interact with each other in a limited manner while evolving. The division between body and control is a useful abstraction to simplify the optimization process. However, the process of developing the body and the brain in nature is intertwined, both of them are ``translated" from the same genome and evolve together. Bongard and Pfeifer \cite{new_int_view} have argued that such body-brain co-evolution is critical toward progress in evolutionary robotics and artificial intelligence.

Pontes-Filho, Walker, Najarro, Nichele and Risi \cite{unified_substrate}\cite{neural_CA} explored this challenge of co-evolving morphology and control as the same entity by using a Neural Cellular Automaton (NCA). By evolving a neural network that defines the rules of a Cellular Automata, they would first grow the body of the robot, and then run the NCA in each cell of the agent, taking into account only local information from neighboring cells to determine their next state. What differs our method from theirs is that we use a different approach to generating the body and controlling, while they use NCA, we use 2 Neural Networks generated by the HyperNEAT algorithm.

While our approach of co-evolving body and brain at the same time may not be suitable for every type of task, as we show in Section \ref{results_section}, we believe that this is an important step to simulate systems with more life-like characteristics.

\section{Preliminaries}\label{preliminaries}
\subsection{Voxel-based Soft Robots}
We use Voxel-based Soft Robots (VSRs) to simulate the co-evolution of morphology and control. As defined by  Medvet, Bartoli, De Lorenzo, and Fidel \cite{adaptable_morph}, ``A Voxel-based Soft Robot is an aggregation of soft cubic blocks, voxels, that can vary their volume in response to a control signal". By this definition, the emitter of these control signals is the controller and the structure made of the blocks is the body.

We chose to use VSRs because of the high degree of freedom it provides when designing a morphology for the agent. At the same time, when designing the robot we can limit its size and place the blocks in discrete positions, making it easy and more efficient to work with. Finally, one additional advantage of VSRs is that they are capable of representing agents with modular design. Although we have not taken advantage of this feature because of the limited size of our robots, this could certainly be expanded in the future.

\subsection{Evogym}
Evogym~\cite{evogym} is a python library that simulate 2D VSRs and is able to evaluate them. We chose to use it to do the simulation because it is 2-dimensional, meaning it consumes less computational resources, and can simulate different types of tasks.

Below, we briefly describe how this library approaches the construction of the robot structure and how we control the robot in it.
\subsubsection{Robot Structure}
\begin{figure}[htbp]
\centerline{\includegraphics[width=\linewidth]{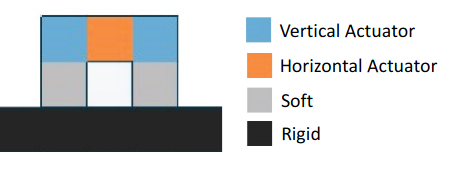}}
\caption{Example of robot structure created over a (2x3) grid in evogym. Image from \cite{evogym}}
\label{evogym_figure}
\end{figure}

Each robot is represented as a matrix of voxels. The value of each voxel is a label corresponding to its type from the set \{Empty, Rigid, Soft, Horizontal Actuator, Vertical Actuator\}. The agent determines the robot body structure only when the environment is initialized. Fig.~\ref{evogym_figure} shows a robot with different voxel types standing in the ground made of rigid tiles.
\subsubsection{Action}
At each time step, the robot’s controller provides an action vector to the environment. Each component of the action vector is associated with an actuator voxel (either horizontal or vertical) of the robot, and it instructs if, and by how much, a voxel should expand/contract.

\subsection{CPPN-NEAT}
A compositional pattern-producing network (CPPN)~\cite{CPPN} is a variation of an Artificial Neural Network (ANN) where its nodes can contain different math functions as activations. Furthermore, both the inputs and outputs of a CPPN operate over a cartesian space in order to generate an image. This is, a coordinate $(x, y)$ in a 2D plane can be used as an input and the output would be the intensity of color at this spot.

CPPN's architecture evolves according to the NEAT algorithm \cite{neat}, meaning that it starts as a minimal ANN with few nodes and connections, but evolves into more complex architectures by using a process similar to natural selection. The CPPN representation is also the same as in NEAT, that is, the ANN is represented by an array that contains a description of the nodes and connections.

In a CPPN, the neural network is the genotype and the produced image is the phenotype. Since the genome does not describe the phenotype directly, CPPNs are considered a form of indirect encoding. 

\subsection{HyperNEAT}
HyperNEAT (Hypercube-based NeuroEvolution of Augmenting Topologies)~\cite{hyperNEAT} is an extension of NEAT and CPPNs that instead of producing an image as the output, produces an ANN.

In hyperNEAT, the genome is a CPPN, but rather than using the coordinates of a single point as input and the intensity of a color as output, the inputs are the coordinates of two points and the output is the weight of the connection between them. By doing this process, connections in an ANN can be defined by the locations of the nodes that they connect; this not only allows the CPPN to build a network whose architecture can exploit the geometry of the problem, but it also allows a genome to create more than one ANN

As defined in the orignal paper \cite{hyperNEAT}, the CPPN 
computes a four-dimensional function $(x_1, y_1, x_2, y_2) = w$, where the first node is at $(x_1, y_1)$ and the second node is at $(x_2, y_2)$. This formalism can return a weight for every connection between every pair of nodes.

The grid of nodes that will be connected is called the substrate. The substrate assigns coordinates for each node in the ANN, and are not limited to be 2-dimensional.

\begin{figure}[htbp]
\centerline{\includegraphics[width=\linewidth]{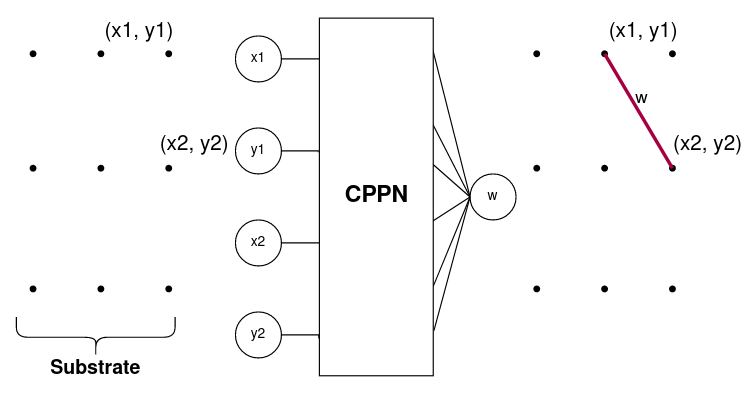}}
\caption{Illustration of the process of querying a connection in hyperNEAT. The two coordinates, $(x_1, y_1)$ and  $(x_2, y_2)$, on the left are queried to the CPPN in the middle. The output of the CPPN is the weight of the connection between coordinates}
\label{hyperneat_fig}
\end{figure}

Fig. \ref{hyperneat_fig} illustrates the process of querying the weight of a connection. In it, the nodes in the substrate are assigned coordinates with the node in the center being the origin. Then, a connection between $(x_1, y_1)$ and  $(x_2, y_2)$ is queried by inputting these coordinates in the CPPN. The output of the CPPN is used as the wight of the red connection that can be seen on the right. In practice, all connections of the desired ANN are queried and the final network is usually a fully connected ANN.

\section{Method Description} \label{methodology}
In this work, we co-evolve body and control of soft robots. Each of these behaviors is governed by a different ANN, meaning that there is one Neural Network responsible for generating the robot body structure and another one responsible for controlling the robot.

In order to generate two ANNs from a single genome we use the HyperNEAT algorithm explained in section~\ref{preliminaries}. By dividing the substrate in two sections and not connecting nodes between these sections, we create two complete separated ANNs.

Since the ANN for the control outputs a 2-dimensional grid, we opted to add an additional dimension to the substrate in order for the CPPN to be able to determine if the nodes it is operating over are in the control or in the body ANN. This means that our method uses a 3D substrate where the controller ANN nodes have negative \textit{z} coordinates, and the morphology ANN nodes have positive \textit{z} coordinates.

\begin{figure}[htbp]
\centerline{\includegraphics[width=\linewidth]{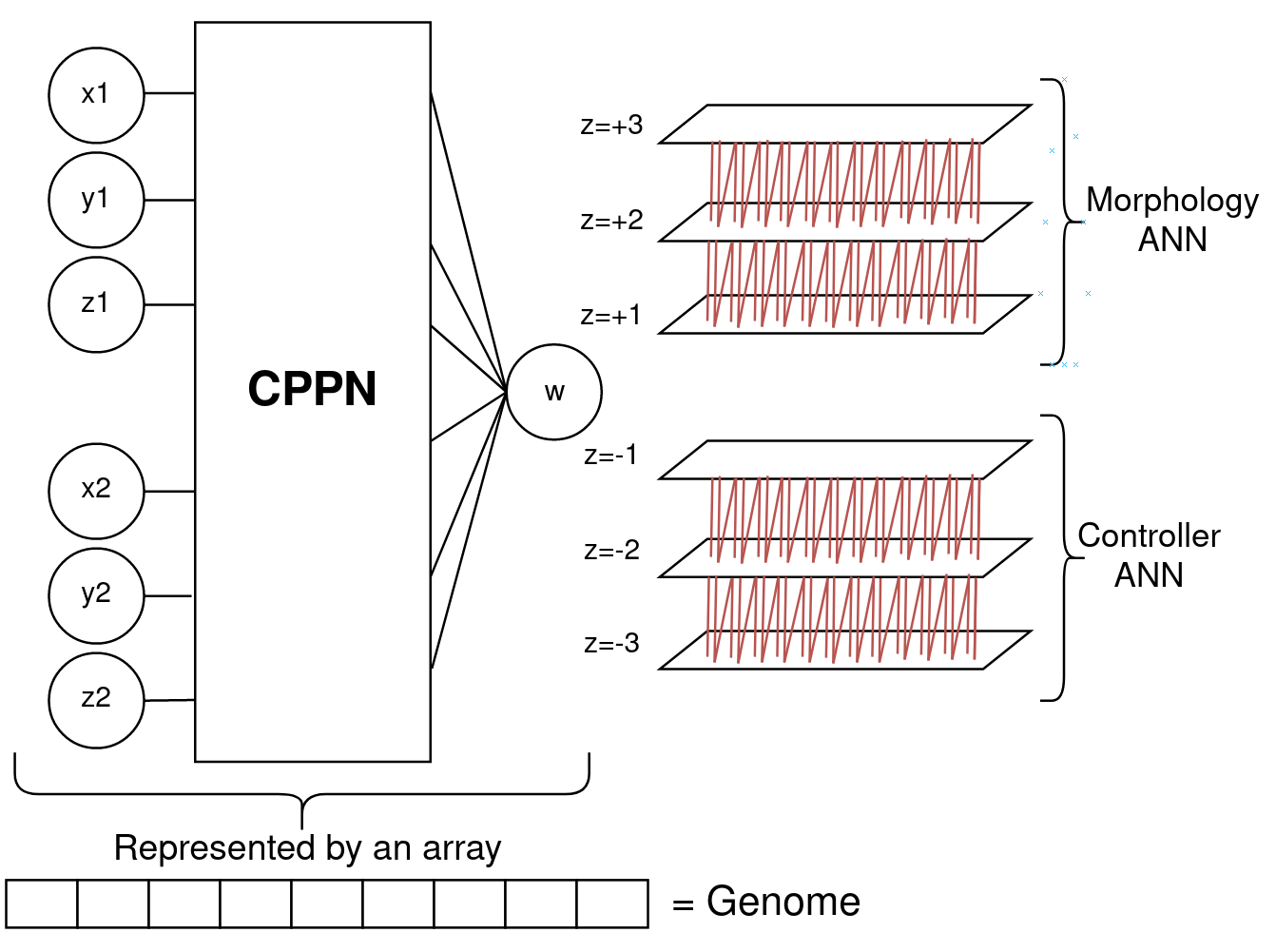}}
\caption{Genome representation. On the left the CPPN is represented by an array and in our method, this array is the genome. On the right there are the two 3-dimensional ANNs generated by our genome}
\label{genome_representation}
\end{figure}

Fig.~\ref{genome_representation}, displays this representation of the genome and how it creates 2 separated ANNs.

One possible implication of our approach of generating two ANNs from a single genome is that any change to the CPPN will most likely affect both control and morphology at once. This could make the coordination between the body and control of the agents harder because agents would not be able to improve one without compromising the other. Furthermore, since the two generated ANNs are responsible for two very different types of process, mutations on the genome will impact each aspect differently. 


The next subsections will explain in detail how we use the ANNs to generate and control the robot and how we perform the evolution and speciation process. 

\subsection{Robot body generation} \label{morphology_creation_description}

\begin{figure}[htbp]
\centerline{\includegraphics[width=\linewidth]{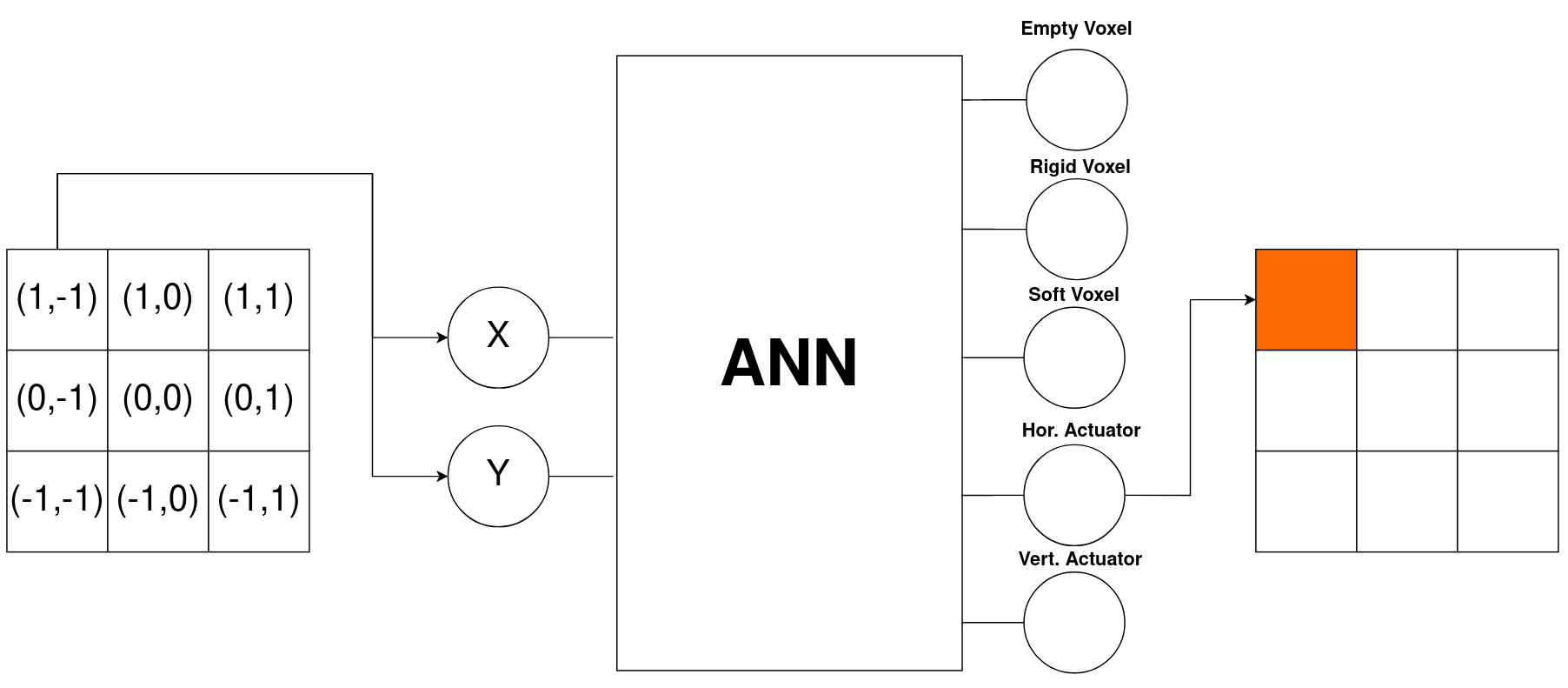}}
\caption{How a robot is constructed from an ANN. In this process, the coordinates of a voxel are input into the ANN and the output of it is the type of voxel the robot will have at this coordinate}
\label{robot_construction}
\end{figure}

We define the ANN that generates the morphology of as follows. The body of the robot is a 2D grid, the origin representing the center of the robot. The ANN has two input neurons corresponding to the $(x,y)$ coordinates of a voxel. The ANN has five output neurons corresponding to four types of voxels and an empty voxel. The voxel at coordinate $x,y$ will be assigned the type corresponding to the output neuron with highest value. Fig.~\ref{robot_construction} exemplifies this processes.

This approach to the generation of the morphology is similar to the one in \cite{evogym} and \cite{unshackling}. However, the difference is that they use a CPPN instead of a ANN and, because of that, are able to use different activation functions. In our method, we didn't use CPPNs simply because the default implementation of hyperNEAT is only capable of generating ANNs.

It is important to note that for a robot to be simulated there are 2 constraints that must be respected: the body has to be connected, and actuators must exist. We do not enforce these constraints when generating the genome. During evaluation, genomes that produces a body that violates these constraints are removed from the population. This is the minimum criterion that any genome must have to be evaluated.

\subsection{Controlling the robot} \label{controller_creation_description}
We define the ANN that controls the robot as follows. The values produced by sensors are the input, and there is one output neuron for each voxel of the robot. The value of these neurons describes if that voxel will expand or contract, and by how much. 

The sensors used are dependent on the task. For example, in the Walker-v0 task the sensors give the speed of the robot and the position of each voxel. For more information, please refer to the Evogym documentation~\cite{evogym}.

\subsection{Evolution process}
In order to evolve the genome population, we use the standard evolutionary algorithm described in the original NEAT paper\cite{neat}. Algorithm \ref{evo_alg} gives a high level overview of the evolutionary process, including how we evaluate the genomes in regards to our environment.

\begin{algorithm}
\caption{Evolutionary Algorithm Overview}\label{evo_alg}
\begin{algorithmic}
\Require environment $env$, \# of generations $n$, \# of simulation steps $s$.
\Ensure final genome population $G$
\State $G \gets \text{GenerateGenomes()}$
\For{$i \gets 1 $ \textbf{to} $n$}
\For{$g \in G$}
\State $B \gets \text{GetBodyFromGenome}(g)$
\State $C \gets \text{GetControllerFromGenome}(g)$
\State $env, obs \gets \text{StartEnv} (env, B)$
\For{$j \gets 1 $ \textbf{to} $s$}
\State action $\gets$ GetActionFromController $(obs, C)$
\State env, obs, reward$\gets$ EnvStep $(env, action)$
\State UpdateGenomeFitness$(g, reward)$
\EndFor
\EndFor
\State $species \gets \text{SpeciationFunction(G)}$
\State $G \gets \text{SelectionFunction}(G, species)$
\State $G \gets \text{ReproductionFunction}(G)$
\EndFor
\end{algorithmic}
\end{algorithm}

\subsection{Speciation}
During the process of evolving the genomes, most new mutations do not increase the fitness, but, after some time, accumulated mutations may lead to a breakthrough and a new better agent. In order to preserve novel structures and allow them to optimize before they are eliminated from the population, the NEAT algorithm, and consequently hyperNEAT, implements a technique called speciation.  This is, agents are divided into species based on a distance derived from the similarity of their genomes. Based on this division, newer or smaller species receive incentives to be preserved and crossover only happens between members of the same species. By doing this, we can have more variety in the genomes and more complex structures can arise over time.

However, during our early experiments, this implementation of speciation would group agents with distinct types of bodies into the same species, and, over time, the representatives with the best fitness of each species would rarely change. We believe this happened because the traditional algorithm considers only the genotypic similarity between genomes. In our algorithm, two similar genomes could generate very different morphologies and controllers since we use an indirect encoding. As a result, agents with distinct body structures would end up in the same species, not only discouraging new morphologies, but also deterring the optimization of a controller for a specific body type.

To alleviate this problem, when calculating the distance that separates the robots into species, we take into consideration both genotypic similarity and a new measure of similarity between bodies (a type of phenotypic distance). This will not only incentive newer morphologies, since new body types will receive an advantage because they will more likely be a new species, but it will allow the crossover to happen only between agents with similar morphologies, facilitating the development of controls for a specific body type.

Equation \eqref{dist} shows how the distance between two genomes is calculated: a sum between the genotypic distance (written as a function called $gDist$) explained in the NEAT algorithm \cite{neat}, and the sum of the difference between every voxel of the genomes (written as the function $b$) multiplied by a constant $v$. 

Equation \eqref{bdist} shows how the distance between two voxels in position $(i, j)$ of different genomes is calculated: if  they are the same type (this includes being empty voxels), the distance is 0; if one of them is empty and the other is not, the difference is 1; and finally, if they are both not empty but are of different types, the distance if 0.5. The reason we have different values for the last 2 cases is because the presence of a voxel where there was none before has more impact on how the robot moves than just simply changing the voxel type.

\begin{equation}
d(g^1, g^2) = gDist(g^1, g^2) + v*\sum_{i=1}^{n}\sum_{j=1}^{n} b(g^1_{ij}, g^2_{ij})
\label{dist}
\end{equation}
\begin{equation}
b(g^1_{ij}, g^2{ij})= 
  \begin{cases}
    0       & \text{$g^1_{ij}$ is the same type of $g^2_{ij}$}\\
    .5      & \text{$g^1_{ij}$ and $g^2_{ij}$ have different type}\\
    1       & \text{one is empty and the other is not}
  \end{cases}
\label{bdist}
\end{equation}

\begin{figure}[htbp]
\centerline{\includegraphics[width=.8\linewidth]{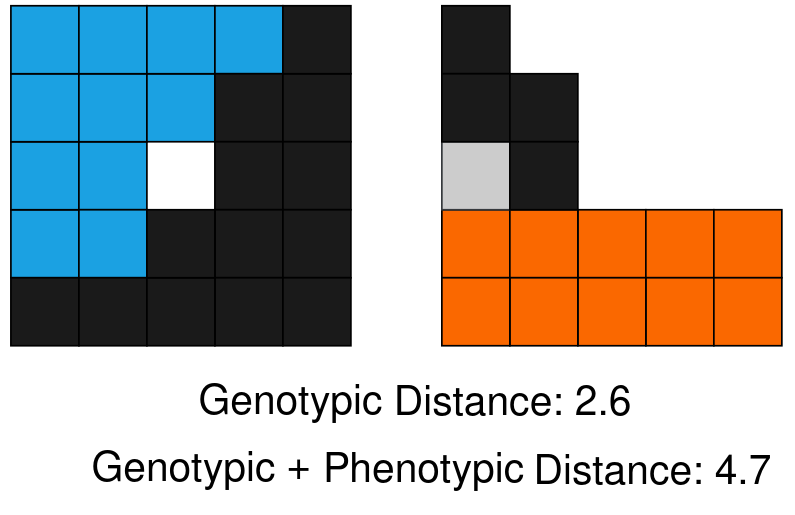}}
\caption{Distance between agents using two different measures. The traditional speciation algorithm considers only the genotypic similarity. Agents that separated by a distance below a certain threshold are considered the same species. Our speciation method takes into consideration both genotypic and phenotypic distances, making it harder for agents with too different body structures to be considered members of the same species.}
\label{distance_example}
\end{figure}

Fig~\ref{distance_example} illustrates how different distance function can impact the speciation. If the threshold for agents to be in the same species is a distance of 3.5, by using only genotypic distance the two agents would end up in the same species even if they have very different bodies. By taking into consideration phenotypic distance too, these agents can be considered different species.

\section{Experiments} \label{experiments}
In order to assess our method of co-evolution of morphology and controller in soft robots, we evaluated it in four different Evogym \cite{evogym} tasks. We then compared the results with other baseline algorithms.

We chose the tasks so that they have have different types of challenges. Two of them are locomotion tasks where the aim is to go the farthest to the right possible, with one being easy because it is in a flat terrain and the other made harder by the presence of obstacles. The third task is related to vertical movement and requires a specific type of agent body. The final task is related to object manipulation and involves handling a box that obeys a simple simulation physics. Below you can find the specification of each task.
\subsection{Tasks}
\subsubsection{Walker-v0}
In this task, the goal is to move as far as possible across a flat terrain in a limited time. The agent has sensors for its speed and position, and its fitness is evaluated by the distance it traverses. We ran this task for 500 steps.

\subsubsection{ObstacleTraverser-v1}
The goal of this task is similar to the previous one. However, the terrain is highly uneven. While the fitness of this task is the same as the first one, the agent now has additional sensors for its own orientation and for the shape of the terrain below itself. We ran this task for 600 timesteps.

\subsubsection{Climber-v2}
In this task, the goal is to climb as high as possible through a narrow stepwise channel. One possible way to achieve this is for the agent to expand and have points of contact with both walls at the same time. The agent fitness is calculated by its final $y$ position and it has sensors for its speed, position, orientation, and the shape of the walls on either side of it. We ran this task for 400 timesteps.

\subsubsection{Thrower-v0}
The goal for this task is for the robot to throw a solid box as far as possible. This box is not directly operated by the controller and the agent needs to move its body to interact with it. The sensors for this task capture the agent's position and speed in addition to the box speed and position. The fitness is evaluated by the final position of the thrown box. We ran this task for 300 timesteps.

\subsection{Baseline Algorithms}
\subsubsection{Nested loop NEAT} 
This algorithm tackles the problem like a two-level optimization problem and uses separate populations to evolve the morphology and the control of the soft robots. In the outer loop, we use NEAT to evolve agents that generates the morphology in the same way as described in section \ref{morphology_creation_description}. Then, in the inner loop, for each unique soft-robot body, we create a new NEAT population that evolves the controller for the robot. 

A more complete description of this approach can be found in the original Evogym paper \cite{evogym}. 
The original algorithm used a policy gradient optimization algorithm called Proximity Policy Optimization (PPO). However, we could not reproduce their results with their available code and instead used the evolutionary algorithm NEAT. 

Please note that since this approach uses two nested loops, the number of evaluations is squared ($n$ morphologies with $n$ controllers each); for that reason we reduced each NEAT population of the controller and morphology to only 12 agents each.

\subsubsection{Direct encoding NEAT} 
\begin{figure}[htbp]
\centerline{\includegraphics[width=.7\linewidth]{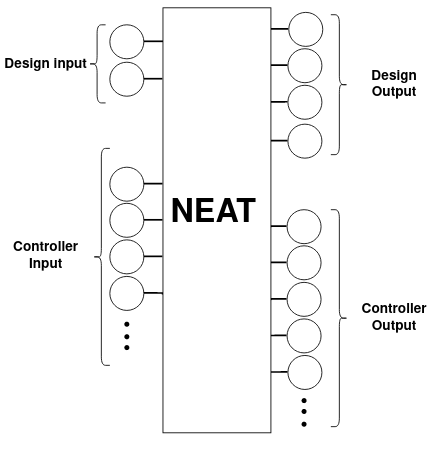}}
\caption{Illustration of how the neural network in the \textit{Direct Encoding NEAT} algorithm is interpreted. The top part is used when specifying the robot morphology and the lower part when controlling the robot.}
\label{singe_genome_neat}
\end{figure}

This algorithm was our first approach to the problem of co-evolving body and control using a single genome. In it, we evolve a single ANN using NEAT. This network is divided in two parts, one for the body and one for control of the robot, with the possibility of having connections between these two parts. Fig~\ref{singe_genome_neat} illustrates this ANN. 

When designing or controlling the robot, we use the same methodology described in sections~\ref{morphology_creation_description} and~\ref{controller_creation_description}. In addition, since there is only one network, the input nodes that are not being used at any given time are given 0 as their value.

This approach is similar to ours in the sense that a single genome contains the description of the body and the brain, however, our method utilizes hyperNEAT to generate two separated ANNs and this algorithm uses only a large neural network where the first half describes the morphology and the second half the controller. 

\subsection{Hyperparameters}
\subsubsection{General Parameters}
Parameters related to the environment specifications and for how long we ran the algorithms can be found on table~\ref{general parameters}. By running our tests in a 12 core CPU, each generation of our method took around 30 seconds. 

\begin{table}[htbp]
\caption{General parameters}
\begin{center}
\begin{tabular}{||c c||} 
 \hline
 Parameter name & Value \\ [0.5ex] 
 \hline\hline
Generations & 250 \\ 
Population size & 128 \\
Robot size & (5x5) \\ 
Repetitions & 5 \\
[1ex] 
 \hline
\end{tabular}
\label{general parameters}
\end{center}
\end{table}

\subsubsection{HyperNEAT hyperparameters}
For our implementation of hyperNEAT, we used the NEAT-python library \cite{neat-python} to evolve and keep track of the CPPN population. The most important hyperparameters used are listed below in table~\ref{neat_param_tab} and their interpretation can be found in the library documentation\footnote{https://neat-python.readthedocs.io/en/latest/config\_file.html}. Note that the input for this CPPN are two 3D coordinates plus a bias, and the output is the weight of the connection between these coordinates. The rest of the parameters were chosen by hand in a way that tries to increase the diversity by having a high chance of mutation. 

\begin{table}[htbp]
\caption{cppn-neat parameters}
\begin{center}
\begin{tabular}{||c c ||} 
 \hline
 Parameter name & Value \\ [0.5ex] 
 \hline\hline
activation\_default & tanh \\
activation\_mutate\_rate & 0.2 \\
activation\_options & sin tanh gauss \\
bias\_mutate\_power & 0.2 \\
bias\_mutate\_rate & 0.8 \\
bias\_replace\_rate & 0.2 \\
conn\_add\_prob & 0.2 \\
conn\_delete\_prob & 0.2 \\
node\_add\_prob & 0.2 \\
node\_delete\_prob & 0.2 \\
weight\_mutate\_power & 0.2 \\
weight\_mutate\_rate & 0.8 \\
weight\_replace\_rate & 0.2 \\
max\_stagnation & 20 \\
species\_elitism & 1 \\
elitism & 2 \\
survival\_threshold & 0.25 \\
min\_species\_size & 4 \\ [1ex] 
 \hline
\end{tabular}
\label{neat_param_tab}
\end{center}
\end{table}

\subsubsection{HyperNEAT substrate}
As discussed in section \ref{methodology}, we utilized a 3-dimensional substrate to generate the ANNs, and, in order for the CPPN to have some information about which ANN it is constructing, the substrate was divided into 2 parts. One can think of each layer of the ANN as a 2D grid in $z$ coordinate, tables \ref{morphology substrate} and \ref{controller substrate} shows how we organized these layers.

\begin{table}[htbp]
\caption{Substrate used to construct the morphology ANN}
\begin{center}
\begin{tabular}{||c c c||} 
 \hline
 Substrate name & Size & layer position\\ [0.5ex] 
 \hline\hline
Input layer & (1x2) & 1 \\ 
Hidden layer & (1x3) & 2 \\
Output layer & (1x5) & 3 \\
[1ex] 
 \hline
\end{tabular}
\label{morphology substrate}
\end{center}
\end{table}

\begin{table}[htbp]
\caption{Substrate used to construct the controller ANN}
\begin{center}
\begin{tabular}{||c c c||} 
 \hline
 Substrate name & Size & layer position\\ [0.5ex] 
 \hline\hline
Input layer & \textit{Task dependant} & -1 \\ 
Hidden layer & (5x5) & -2 \\
Output layer & (5x5) & -3 \\
[1ex] 
 \hline
\end{tabular}
\label{controller substrate}
\end{center}
\end{table}

\subsection{Reproducibility}
For reproducibility purposes, all the code and experimental scripts are available online\footnote{https://github.com/fhtanaka/SGR}.

\section{Results} \label{results_section}
This section introduces the results of the experiments described in Section~\ref{experiments}.

\begin{figure}[htbp]
    \centering
    \includegraphics[width=0.9\linewidth, keepaspectratio]{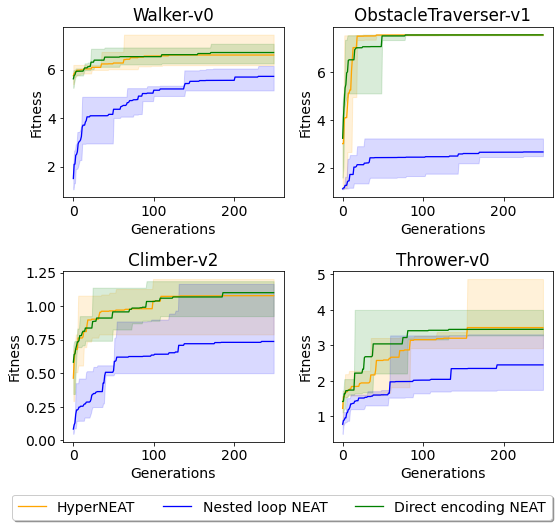}
    \caption{Performance comparison among the algorithms. We plot the best performance of robots that each algorithm has evolved in each generation. All the curves are averaged over 5 different runs, and the variance is shown as the shaded region.}
    \label{fit_results}
\end{figure}

The graphs in Fig.~\ref{fit_results} show how the algorithms performed in each task in regards to its fitness over the generations. It is possible to see that, from the three methods, the Nested loop NEAT performed worst in all tasks. At the same time, our approach, named HyperNEAT in the image, had very similar results to Direct encoding NEAT and performed well in all but the Climber-v2 task. 

\begin{figure}[htbp]
\centerline{\includegraphics[width=\linewidth]{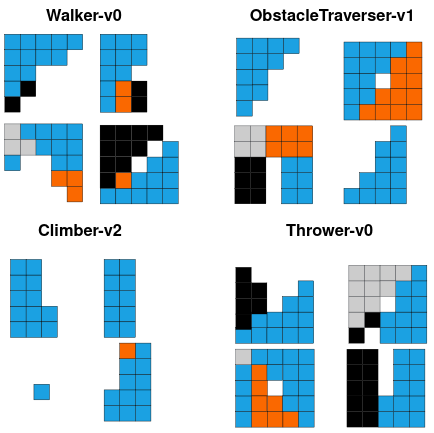}}
\caption{Examples of morphologies created by our method in different tasks. The colors of the voxels represent their types as follows: black are rigid voxels, gray are soft, orange are horizontal actuators and blue are vertical actuators}
\label{morph_ex}
\end{figure}

\begin{figure*}[htb]
    \centering
    \includegraphics[width=0.8\textwidth]{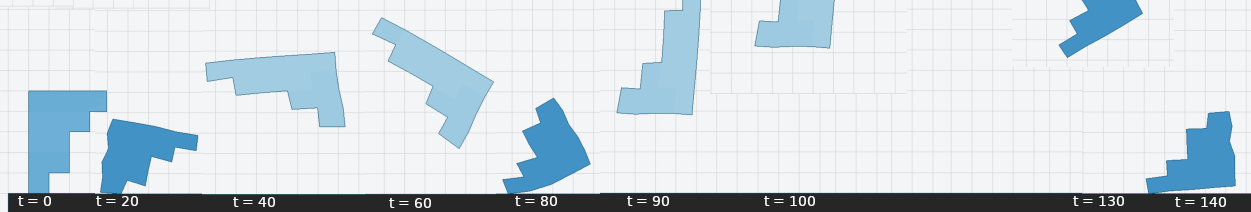}
    \caption{Illustration of an unconventional way that the controller moves the robot. In this image, the voxel gets darker when contracting and lighter when expanding. This agent moves by doing front-flips and is able to achieve high speeds.}
    \label{timelapse}
\end{figure*}

We believe that the main reason for the HyperNEAT and Direct encoding NEAT better performance is that, by having a single genome, they had a larger sample of robot bodies at any time. Since the Nested NEAT method has to divide its agents between the population that evolves the control and the one that evolves the morphology, there are fewer body structures evaluated. This also explains why there is such a large gap between the starting fitness between the methods.

One of our initial concerns when developing our method was that every mutation on the genome would affect the controller and design at the same time and thus would be destructive by preventing the genome to develop any coordination between the body and brain. However, this was not the case; the hyperNEAT method performed similarly to the Direct encoding NEAT algorithm, an algorithm where, most of the time, a single mutation does not affect control and design at the same time. This is not to say that there were no destructive mutations, but if there were, they were eliminated from the population during the evolution process. The hyperNEAT method was even capable to evolve robots with complex controls that presented some unexpected movements as illustrated in Fig.~\ref{timelapse}, where the robot moved forward at high speed by doing front-flips. 

Fig.~\ref{morph_ex} displays some examples of morphologies generated by the hyperNEAT algorithm in different runs. It is possible to see that, even without multiple activation functions, our method was capable of producing varied and mostly symmetric body structures, many times presenting areas with the same voxel type. The reason for this type of regularity is that, when designing the body, we use a method similar to a CPPN; meaning that the input for the ANN that defined the morphology was the coordinate of the cell in a grid, and cells on opposites sides give similar values, just changing if they are positive or negative. 

One of the advantages of having these regular body structures is that the voxels can work in a coordinated manner to do a specific type of movement. For example, all blue voxels in a structure can contract and expand at the same time to function in a way similar to a muscle. This type of coordination is what allowed many of the agents to move by jumping.

\subsection{Behavior analyze per task}
The first task we evaluated, Walker-v0, is a simple task where the goal is to move on a flat surface. Since our initial population wasn't so small, some agents were able to move and achieve good fitness already from the start. The way most agents moved was by performing big jumps, by doing this, they would cover more ground in a shorter time span compared to walking. Fig.~\ref{timelapse} illustrates how one of the best-performing agents would jump.

The following task, ObstacleTraverser-v1, is a harder version of the previous task because the terrain is uneven. This difficulty was reflected in the fitness of the starting population that wasn't able to move very far. However, the agents evolved fast to have the same behavior as the previous task and move by jumping. By performing big jumps, many agents completely went over most of the uneven ground and when they landed, they just jumped again. Since the terrain was the same in all runs and generations, agents were able to exploit the simulation and perform jumps that would always land in the same place and then jump again from there.

The task where agents performed the worst was the Climber-v2, and by looking at the generated morphologies it is possible to interpret why. All the generated structures are too thin and were not able to cling to the walls. They achieved their fitness by simply jumping high but got stuck in this behavior as a mlocal maximun. This exemplifies one of the drawbacks of our method, some environments favor irregularities in the body and our algorithm could not produce that. 

The final task, Thrower-v0, is a manipulation task where the agents would throw a box the farthest distance possible. Agents for this task evolved at a more linear rate and presented some very interesting bodies. Some of them were basically squares that would tilt forward and then expand to shoot the box. The best-performing agents however, had a "two towers" body structure like the one in the bottom right of Fig~\ref{morphology substrate}; they would use the right tower to propel the box while the left tower would be used to aim. 

\section{Conclusion}
In this study we introduced a method for co-evolving the morphology and controller of a soft robot in the same process.
The method uses the hyperNEAT algorithm to generate two ANNs, one that designs the robot body structure and one that controls the robot in the environment. Our method puts both the ``brain" and the ``body" of the agent in a single genome, similar to what happens in nature.

We evaluated our method by executing it in four different tasks and observed that the generated agents were able to adapt well to almost all of them. The body structures generated presented symmetry and their controllers were able to exploit regular structures in order to achieve their goals. One of our concerns, that every mutation would always affect both controller and morphology and thus not allow coordination between controller and body, did not seem to affect the agents, at least not in a way that prevented them from evolving.

Overall, this work showed that it is possible to evolve both morphology and controller at the same time. This does not invalidate any other previous method of co-evolution that treats body structure and control as separate procedures, but can be useful to reproduce and study natural evolution processes. Going forward, there are many directions to expand this research, one particular of particular interest would be studying new ways to define a robot structure from an ANN, this could allow the hyperNEAT algorithm to take advantage of the geometry of the problem and even generate bodies that are may be less regular and thus solve specific tasks.

\section*{Acknowledgements}

This work was supported by JSPS grant 22K11918.

\end{document}